\documentclass[letterpaper, 10 pt, conference]{ieeeconf}  

\IEEEoverridecommandlockouts                              

\usepackage{times,amsmath,amsfonts,stmaryrd,multirow,subfigure,epsfig,enumerate}
\usepackage{url}
\usepackage{flushend}
\usepackage{cite}
\usepackage{array}
\usepackage{color}
\usepackage{pdfsync}
\usepackage{textcomp}
\usepackage{graphicx}
\usepackage{algorithm}
\usepackage{algcompatible}
\usepackage{textcomp}
\usepackage{xfrac}

\title{\Large \bf
Generative Autoregressive Networks for 3D Dancing Move Synthesis from Music
}%
\author{Hyemin Ahn$^{1*}$, Jaehun Kim$^{2*}$, Kihyun Kim$^{1*}$, and Songhwai Oh$^{1}$
\thanks{$^{1}$H. Ahn, K. Kim, and S. Oh are with the Department of Electrical and Computer Engineering and ASRI, Seoul National University, Seoul, 08826, Korea (e-mail: hyemin.ahn@rllab.snu.ac.kr, \{hahakhkim, songhwai\}@snu.ac.kr).}%
\thanks{$^{2}$J. Kim is with the Delft University of Technology, Delft, The Netherlands (e-mail: j.h.kim@tudelft.nl).}%
\thanks{$^{*}$These authors are equally contributed to this work.}%
}

\begin{document}

\maketitle
\thispagestyle{empty}
\pagestyle{empty}

\begin{abstract}
This paper proposes a framework which is able to generate a sequence of three-dimensional human dance poses for a given music.
The proposed framework consists of three components:
a music feature encoder, a pose generator,
and a music genre classifier.
We focus on integrating these components
for generating a realistic 3D human dancing move from music, 
which can be applied to artificial agents and humanoid robots.
The trained dance pose generator, which is a generative autoregressive model, is able to synthesize a dance sequence longer than 5,000 pose frames. 
Experimental results of generated dance sequences from various songs show how the proposed method generates human-like dancing move to a given music.
In addition, a generated 3D dance sequence is applied to a humanoid robot, showing that the proposed framework can make a robot to dance just by listening to music. 
\end{abstract}

\section{Introduction}

Dance is one of the most important form of performing arts that having been emerged in all known cultures.
As one of the specific subcategory of under theatrical dance,
choreography associated with music is also one of the most popular forms
that have usually been designed and physically performed by professional choreographers.
Recently, there has been a number of new attempts to profit commercially with the dancing character \cite{kda}.
Specifically, a vocal group, whose members are all fictional and in 3D animated characters, succeeded attracting substantial attention by their music and choreography.
This choreography has been obtained by using an expensive motion capture equipment with professional artists, which is a time consuming and costly task.
However, since the obtained choreography can only be used for a single song,
the motion capture process needs to be repeated to create a choreography for another music,
unless the agent mimics the human ability to generate dances.
To overcome this limitation, this paper proposes a novel framework
for automatic dance generation 
which can synthesize a sequence of 3D dancing moves from music.

%
%
%

There have been several attempts related to 3D human motion modeling.
There exist studies related to 3D joint sequence prediction
for the character motion synthesizing
such as human locomotion generation \cite{holden2016deep, pavllo2019modeling}.
However, these are for generating a motion itself,
while our objective is generating a motion
dependent to the specific source sequence.
Related to motion generation dependent on a specific source sequence, \cite{ahn2018text2action} has proposed a network for generating a motion of 3D upper body skeleton according to the given language sentence explaining a specific human behavior. 
When the input condition sequence is music, 
\cite{shlizerman2018audio} has succeeded in producing a upper body motion of an avatar playing a violin or a piano 
when a classical music has been provided as an input.
Our goal is similar to \cite{shlizerman2018audio}
in that we regard music as input, 
however, we suggest a methodology
for a more challenging task of generating the appropriate 3D full-body human dance sequence to the provided music.

Even for a human, generating a dance sequence 
is a challenging task, requiring talents and experiences. 
One needs to understand a rhythmic feature and mood of a given music,
and create a motion sequence that can meet the aesthetic criteria
constrained by the musical content.
In this paper, we aim at enabling an intelligent system
to be capable of such process using a data-driven approach, namely deep neural networks. 
The proposed framework is composed as follows:
a music feature encoder which generates a feature vector containing the information of a given music, 
a set of pose generators which is trained based on each genre dataset, such as cha-cha, rumba, tango, and waltz dances \cite{relate_1}, 
and a music genre classifier selecting which pose generator to use based on the classified genre of a given music.
\begin{figure}[t]
\begin{center}
    \includegraphics[width=\linewidth]{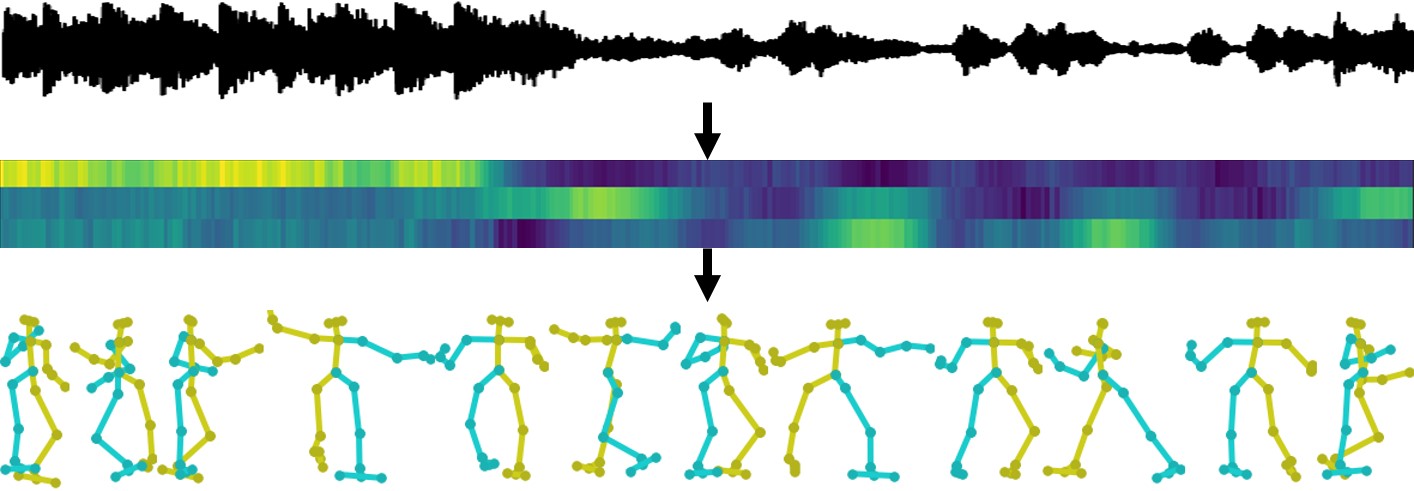}
\end{center}
\caption{An example of a generated dance sequence of $12$ seconds to the song of \textit{Call me maybe} by \textit{Carly Rae Jepsen}. From the top, we have an input audio signal, a generated audio feature, and a generated 3D dancing moves. Note that the dimension of the plotted audio feature has been reduced using principal component analysis (PCA) for better visualization.}
\label{fig:init_fig}
\end{figure}

The proposed pose generator is a generative autoregressive model which does not employ the structure of recurrent neural networks (RNNs) \cite{lstm}.
Our generator predicts the next pose frame based on the music feature and the set of poses generated during a certain period of time in the past.
It consists of a dilated causal convolution layers, which has been shown to be effective when generating a sequence of a large number of frames \cite{bai2018empirical, oord2016wavenet}, and fully connected (FC) layers. 
In Section~\ref{sec:exp}, 
it is shown that our model is efficient in that it can provide better qualitative performance with fewer parameters than RNN-based models \cite{relate_1}.

Experimental results show examples of generated dance sequences, each sequence with more than 5,000 pose frames.
Comparison with other baseline methods \cite{relate_1} demonstrates
that the proposed method is more effective in synthesizing dynamic movements according to the given music.
In addition, we apply the generated pose sequence to a humanoid robot NAO, so that the robot can also dance to the given music.

The remainder of the paper is structured as follows.
The related works of dance generation are introduced in Section~\ref{sec:relate}.
Section~\ref{sec:method} describes the structure of proposed framework, which consists of music feature encoder, pose generator, and a music genre classifier. 
Section~\ref{sec:exp} shows the results of generated dance sequences from various songs, and compares the proposed method with other baseline methods. 
The demonstration of the proposed method using a humanoid robot NAO is also provided. 


\section{Related Work} \label{sec:relate}

There have been several approaches 
for generating 3D dancing moves of a human
\cite{relate_4, relate_1, DBLP:journals/corr/abs-1811-00818}.
In \cite{relate_4}, the authors proposed a model that can randomly generate a choreography based on the dataset of movements of dancing humans recorded by a commercial sensor array.\footnote{Microsoft Kinect}
Although the generated dance sequence is a potential guidance for the choreography design process of a human, it does not utilize the audio information as input
so the resulting dance has no correlation with the music.
On the other hand, a model proposed in \cite{relate_1} is a recurrent neural network (RNN) based auto-encoder
which learns the relationship between music and pose features. 
The model encodes the music feature,
which consists of $16$-dimensional pre-extracted acoustic features and three temporal indices,
into a latent vector in order to predict $63$-dimensional pose features. 

The acoustic feature used in this work is based on a set of hand-crafted features that are popularly used in music and audio domain, and the temporal indices are related to each audio frame's location and beat per minute (BPM) information.
The pose feature is a set of values of joint positions
which are relative to the center of mass of skeleton. 
The method proposed in \cite{relate_1} has a similar concept with ours
in terms of learning the relationship between audio feature and pose.
To show the effectiveness of our proposed method, 
we have analyzed the dance generation results in terms of 
the number of parameters of the network. 
%
%
Experimental results in Section~\ref{sec:exp_compare}
show that our model can synthesize
more realistic dances than \cite{relate_1} even with fewer network parameters.
%

%
%
%
Our approach shares a similar concept with \cite{DBLP:journals/corr/abs-1811-00818}
as it predicts next pose
based on the audio feature and pose data from a certain past period.
However, unlike the moving 3D human shape generated by our proposed methodology,
the pose skeleton generated by \cite{DBLP:journals/corr/abs-1811-00818}
does not provide realistic dancing moves,
since the position of the root joint is always fixed at a specific position and pose data lies on the two-dimensional coordinate.

\begin{figure}[t]
\begin{center}
    \includegraphics[width=\linewidth]{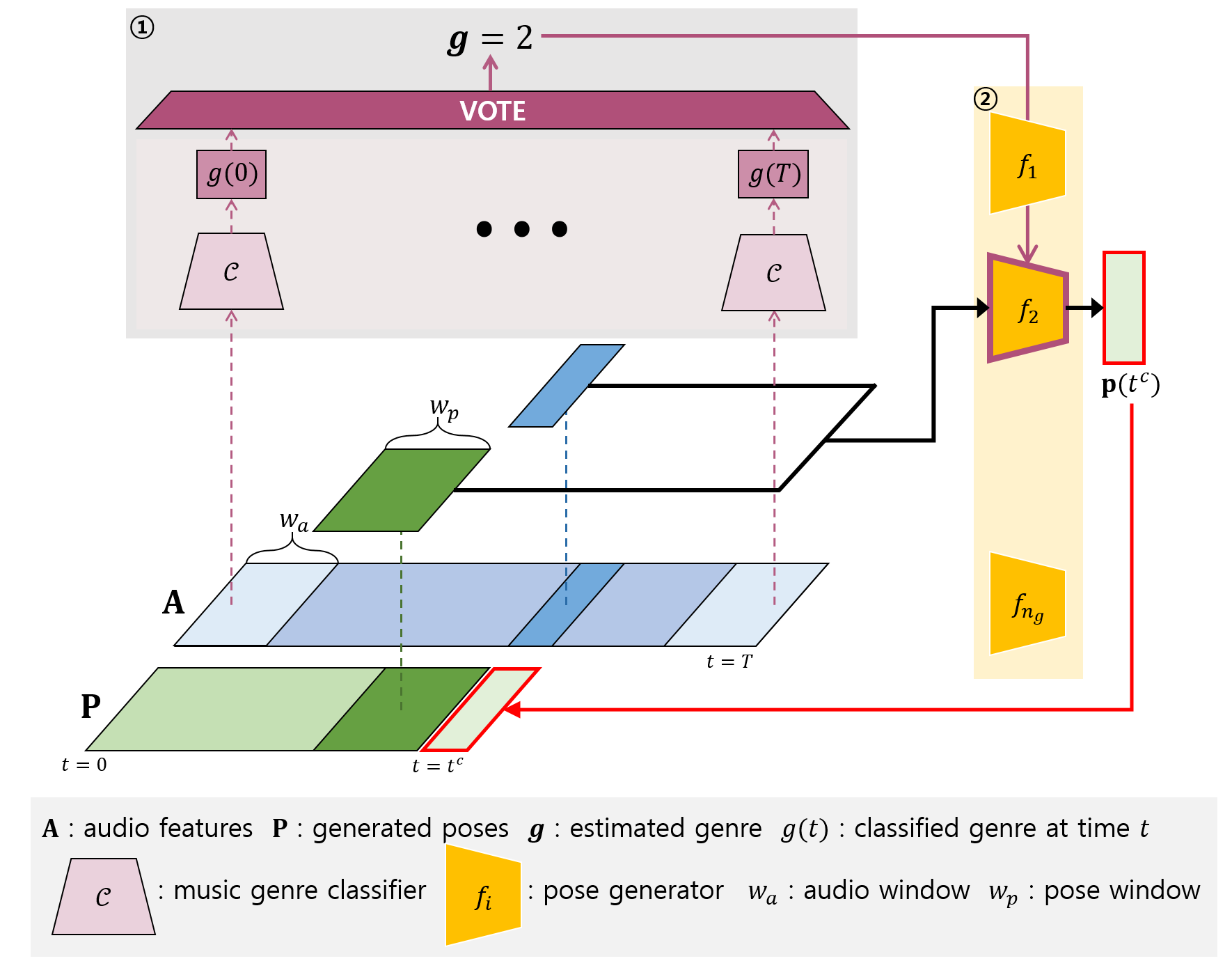}
\end{center}
\caption{An overview of the proposed system. It first determines the genre of the input music, chooses a pose generator for the determined genre, and generates a pose based on the selected model.}
\label{fig:overview}
\end{figure}

\section{Methodology} \label{sec:method}
\subsection{Overall Structure}
We focus on building a machine learning-based framework for generating a 3D pose sequence of a dancing human when music has been given as an input.
Figure~\ref{fig:overview} shows an overview of the proposed system. 
The proposed dance generation system is composed of a music feature encoder, a pose generator, and a music genre classifier. %
In order to synthesize a dance movement tailored to the given music, 
the proposed framework works in the following order.

\begin{enumerate}
  \item Convert the input music into a set of audio feature vectors.
  \item Determine the genre of the music by considering the entire audio feature vector.
  \item Choose the one pose generator according to the identified genre.
  \item Based on the chosen model, estimate the pose for all frames.
\end{enumerate}

After the music feature encoder generates a set of audio feature vectors $\mathbf{A} =\{\mathbf{a}(t)\}_{t=0}^T$, where $T$ denotes the total number of time frames of the given audio,
a music genre classifier $\mathcal{C}$ estimates the genre of the input music at time $t$
by considering a set of audio features such that $\mathcal{C}\big(\{\mathbf{a}(t-{w_a}/2), \ldots, \mathbf{a}(t+{w_a}/2)\}\big)=g(t)$.
Here, 
$w_a$ denotes the length of audio window 
and $g(t)\in \{1,\ldots, n_g \}$ where $n_g$ denotes the number of genres.
Given $\boldsymbol{g}\in\{1,\ldots n_g\}$, which denotes the estimated genre considering $\{g(t)\}_{t=0}^{T}$ ($\boldsymbol{g}=2$ in Figure~\ref{fig:overview}),
the proposed system chooses the $\boldsymbol{g}$-th pose generator $f_{\boldsymbol{g}}$ for generating dance poses for all audio frames.

For generating a pose vector at time $t$, 
the pose generator $f_{\boldsymbol{g}}$ takes the audio feature $\mathbf{a}(t)$,
and a set of previously generated pose vectors $\{\mathbf{p}(t-w_p-1), \ldots, \mathbf{p}(t-1)\}$,
where $w_p$ denotes the length of the pose window.
Since our model is autoregressive, 
the generated 3D pose vector $\mathbf{p}(t)$ is appended 
to the set of generated poses such that $\mathbf{P}=\{\mathbf{p}(0),\ldots,\mathbf{p}(t)\}$, 
and is used for generating the next pose at $t+1$.

\subsection{Music Feature Encoder} \label{sec:music}
To accelerate the learning process, transfer learning is considered for the lower layers where the music signal is encoded. Specifically, we trained a neural network that estimates the beat per minute (BPM) of the given audio signal, which reported as an effective source task to be transferred to the task of which temporal dynamic is important feature such as dance music genre classification~\cite{Kim2019}. After pre-training, we employed the lower layers of the BPM estimation model as the music encoder layers of the proposed system.

\subsubsection{Dataset and Preprocessing}

Training of a BPM estimator model requires a set data points that is a tuple of the music audio signal $x$ and the corresponding BPM value $y_{BPM}$ of given music audio, which leads to the dataset $\mathcal{D}_{BPM} = \{(x^{(i)}, y_{BPM}^{(i)})| i \in \{1...I\}\}$. We exploited the Million Song Dataset (MSD)~\cite{Bertin-MahieuxEWL11} that provides various metadata on commercial songs. MSD includes the BPM information for the all entries, which can be directly used for our learning setup.\footnote{Note, that the BPM provided from MSD is computed from BPM estimation algorithm, which well known for its frequent octave error where the erroneous estimates are the integer multiple of its ground-truth value.} Further, we standardized the BPM values to regularize the loss function.

As for the audio signal, we used the snippets of the 30-seconds music preview at the sampling rate of 22,050 Hz, which are collected with \texttt{7digital} API\footnote{http://docs.7digital.com/}, to extract a low-level feature of audio which is served as the input data. More specifically, we applied the $\mu$-law encoding and decoding on the raw audio~\cite{OordDZSVGKSK16}, where we choose $255$ as the quantization resolution. The decoding process is simply applying the inverse function of encoding process on the quantized data. While the quantization compresses the original audio samples efficiently, it is reported that the encoded representation can be still effectively used as the data source of the learning process~\cite{OordDZSVGKSK16}. We applied the encoding for storing and serving the data for efficiently employ the data at scale, and applied the decoding on the fly when the data is input to the models for the optimization.




Finally, we employed 200,000 / 10,000 / 13,673 tracks for the training, validation, and the testing of the model, respectively.

\subsubsection{Network Architecture}

\begin{figure}[t]
    \centering
    \includegraphics[width=\linewidth]{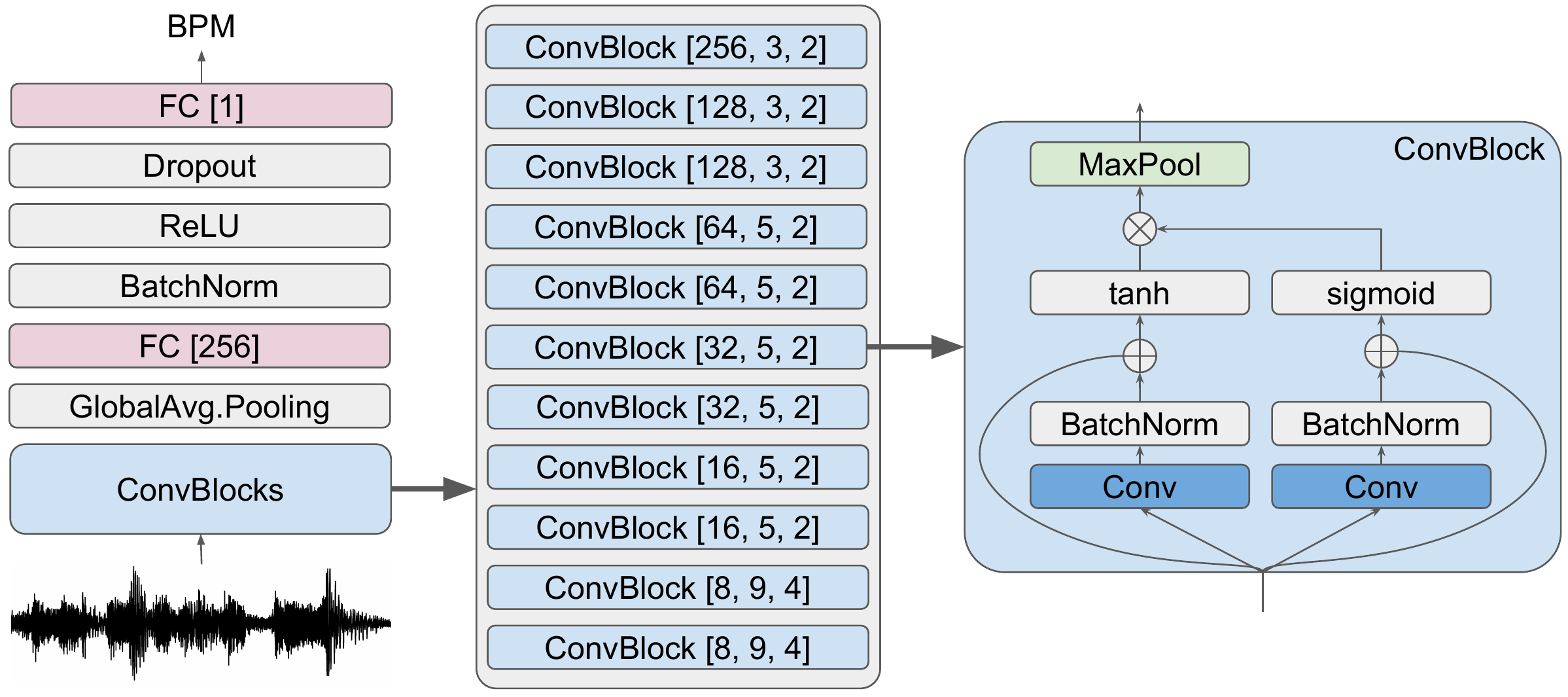}
    \caption{The general architecture of BPM estimator. The network consists of a series of \texttt{ConvBlock} that contains convolution, pooling and other operations. A raw audio signal encoded by these blocks serves to the following regression layers. The numbers inside the brackets refer the relevant hyper parameter regarding each block: For FC layers, it indicates the number of output units. As for \texttt{ConvBlock}, the triplet refers the number of output kernels, the size of each kernel, the ratio of subsampling, respectively. The rightmost block illustrates the inner structure of each \texttt{ConvBlocks}.}
    \label{fig:bpm_net_arch}
\end{figure}

As for the network architecture, we employed the 1-dimensional convolutional neural network (CNN) architecture introduced in~\cite{KimLN18}, which is 1-dimensional analogous of the \textit{VGG-like} networks~\cite{Simonyan2014VeryRecognition}. It consists of cascading convolution and pooling operations, which is shown as effective not only on the image recognition tasks, but also the audio and music related tasks~\cite{Kim2019}. In addition to its base architecture, we added several additional components that are reported as effective on audio and music related tasks~\cite{OordDZSVGKSK16, KimLN18}: batch normalization~\cite{Ioffe}, residual connection~\cite{he2016deep}, and dropout~\cite{Srivastava2014Dropout:Overfitting}. Finally, we choose the gated \textit{tanh} function as the core non-linearity~\cite{OordDZSVGKSK16} for convolution blocks and rectified linear unit (ReLU)~\cite{Nair2010RectifiedMachines} for later fully connected (FC) layers. The general overview of the architecture can be found in Figure~\ref{fig:bpm_net_arch}.

\subsubsection{Training}

The training of the network is achieved by minimizing the mean squared error between the ground truth $y_{BPM}$ and the estimation $ f_{BPM}(x;\Theta)$ as follows:

\begin{equation}
    \mathcal{L}(\mathcal{D}_{BPM}, \Theta) = \mathbb{E}_{(x, y_{BPM}) \sim \mathcal{D}_{BPM}}[(y_{BPM} - f_{BPM}(x;\Theta))^2]
    \label{eq:BPM_loss}
\end{equation}
where $\Theta$ is the set of parameters of BPM estimator $f_{BPM}$ and input data $x \in \mathbb{R}^{44100}$ is randomly selected from the training set at every iteration, and also the randomly cropped 2 seconds chunk (44100 samples) out of the original 30 seconds, which is common in the music and audio task domain~\cite{Kim2019, KimLN18}.
We applied the Adam optimizer for the robust optimization~\cite{Kingma2014Adam:Optimization} in which the parameters are updated in total 600 epochs. To adapt the training at the scale we tested, we applied the mini-batch stochastic gradient with the batch size $M=128$.

\subsubsection{Transfer}

After pre-training, we transferred the first 10 \texttt{ConvBlocks} $\hat{f}_{BPM}$ to the main system as the music recognition pipeline that encodes the input audio into the feature vector $\mathbf{a}\in\mathbb{R}^{128}$ as follows:

\begin{equation}
    \mathbf{a} = w \cdot \Psi
    \label{eq:feature_post_proc}
\end{equation}
where $\Psi=\hat{f}_{BPM}(x; \hat{\Theta}) \in\mathbb{R}^{10\times128}$ is the output of the 10th \texttt{ConvBlocks} for given signal $x$ temporally centered on the MoCap frame where the pose frame is located. The first dimension of the output tensor is the temporal dimension, which still is required to be reduced to represent the music input for each pose frame. We applied the normalized Gaussian window 
$w(j) = \sfrac{e^{-\frac{1}{2}j^{2}}}{\sum_{k=1}^{10} e^{-\frac{1}{2}k^{2}}}, \quad j=\{1...10\}$
for pooling. It eventually summarizes the given temporal feature exponentially more weighted on the temporally close to the pose frame.


\subsection{Pose Generator}
\subsubsection{Dataset and Preprocessing}
For training our pose generator, 
we use a dataset from in \cite{relate_1},
which consists of four types of 3D human dance poses synchronized with each genre of songs.
This dataset provides 3D motion data which have been obtained from a motion capture system with 25 fps, however, the center of mass of the human pose is always fixed in the middle so that the human always remains in the center.
Since the movement of the center of mass is one of the crucial components of dance, we have preprocessed the motion data again so that a human can move along footsteps.

After the preprocessing, we convert the dataset into a format which has been suggested in~\cite{holden2016deep}.
According to~\cite{holden2016deep}, our pose dataset consists of 63-dimensional 3D joint positions defined in the body's local coordinate system, 
the forward direction of the body (3-dimension), 
the global velocity of the body in the floor plane (3-dimension), and rotational velocity of the body around the vertical axis (1-dimension), and foot contact labels of left/right heel or toe (4-dimension) such that the total dimension of a pose vector is $\mathbf{p} \in \mathbb{R}^{74}$.
For the description of the foot contact label, 
let $h_{f}$ denote the height of the foot away from the floor plane. 
Regarding this, we use $e^{-\beta_f h_f}$ as the value of foot contact label with $\beta=10$. 
After the conversion, a set of total pose vectors $\mathbf{P}$ in training dataset are normalized with its mean and standard deviation values. 

\begin{figure}[t]
\begin{center}
    \includegraphics[width=\linewidth]{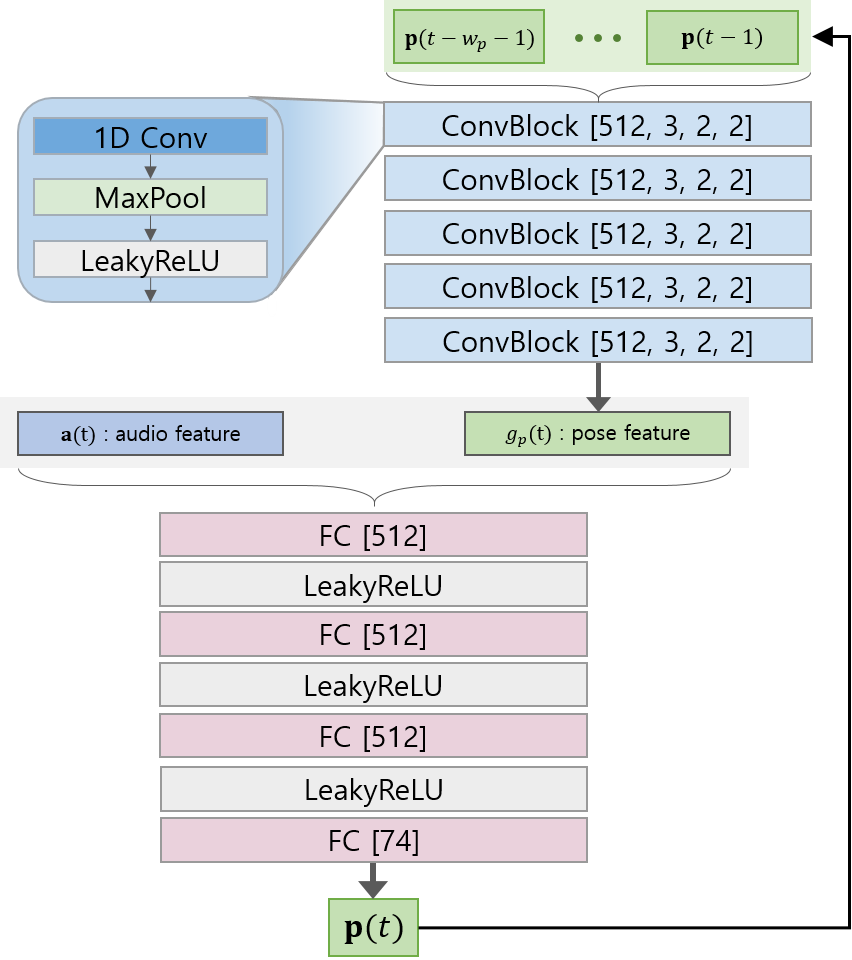}
\end{center}
\caption{The architecture of a proposed pose generator. The numbers inside the brackets refer the relevant hyperparameters regarding each block: for fully connected (FC) layers, it indicates the number of output units, for \texttt{ConvBlock}, the quadruplet indicates the number of output kernels, the size of each kernel, the ratio of subsampling in max-pooling, the stride value in max-pooling.}
\label{fig:generator}
\end{figure}

\subsubsection{Network Architecture}
Let $\mathbf{a}(t)$ denote an extracted audio feature at time $t$, and $\mathbf{p}(t)$ denote a generated pose vector at time $t$.
When a person dances,
one considers the characteristics of the current music and 
how s/he has been moving.
Therefore, in order to generate $\mathbf{p}(t)$, 
the proposed pose generator considers $\mathbf{a}(t)$ and a set of pose vectors that have been generated for a certain period frames, 
such that $\mathbf{P}(t-1;w_p)=\{\mathbf{p}(t-w_p-1),\ldots,\mathbf{p}(t-1)\}$, where $w_p$ denotes the window length of poses and $w_p=32$ has been used in our experiment.
In the test phase, $\mathbf{P}(t-1;w_p)$ is filled up with the zero-valued vector, which is the mean value of $\mathbf{P}$ after the normalization.

\begin{figure}[t]
\begin{center}
    \includegraphics[width=\linewidth]{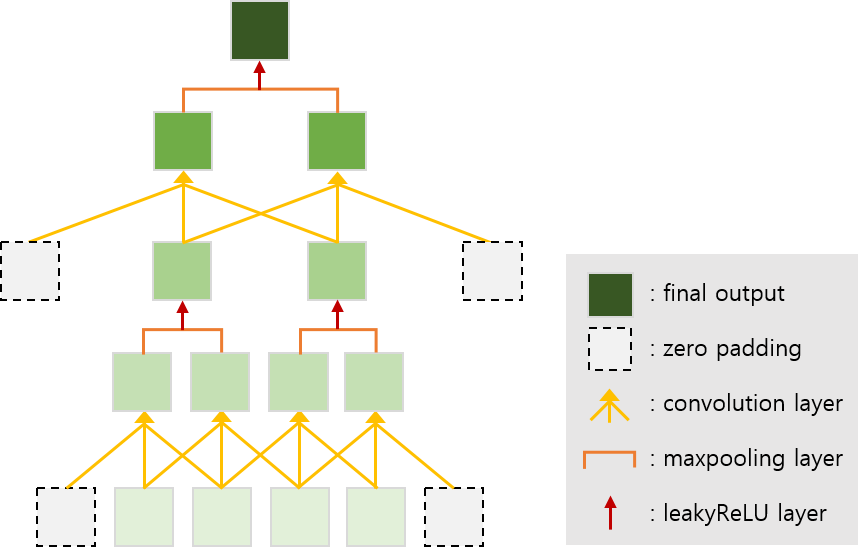}
\end{center}
\caption{The architecture of \texttt{ConvBlock} used in proposed pose generator and music genre classifier.}
\label{fig:convblock}
\end{figure}

As shown in Figure~\ref{fig:generator}, 
the proposed network encodes a feature $g_p(t)$ from $\mathbf{P}(t-1;w_p)$ based on the series of $\texttt{ConvBlocks}$. 
Each \texttt{ConvBlock} consists of 1D convolutional layer, max-pooling layer, and leaky ReLU layer as shown in Figure~\ref{fig:generator} and~\ref{fig:convblock}.
This dilated convolution structure resembles~\cite{bai2018empirical} in that it is causal and the receptive field can be larger with fewer parameters and layers. 
In our proposed network, the length of the receptive field decreases to the half after passing each \texttt{ConvBlock} as shown in Figure~\ref{fig:convblock}.
Note that \texttt{ConvBlock} is also used in our music genre classifier.

After concatenating the generated pose feature vector $g_p(t)$ with the audio feature vector $\mathbf{a}(t)$, the proposed pose generator synthesizes the output pose $\mathbf{p}(t)$ after passing a series of fully connected layers and leaky ReLU layers.
Since our model is autoregressive, the generated $\mathbf{p}(t)$ is appended to $\mathbf{P}(t-1;w_p)$ so that $\mathbf{P}(t;w_p)$ can be used for generating the next pose vector $\mathbf{p}(t+1)$.

\subsubsection{Training}
When training a model which generates a sequence, 
one can follow the strategy called as teacher-forcing method \cite{teacher_forcing},
which gives the \textit{ground truth} $t$-th output to the model as an input when generating $(t+1)$-th output. 
This method has the advantage that the model can quickly learn how to generate the data from the ground truth dataset. 
However, when a model is autoregressive, it can make the model vulnerable to its own prediction error, when a model is exposed by its generation results in a test phase.
Regarding this, one can use a strategy called as student-forcing method in a training phase,
which gives the \textit{generated} $t$-th output to the model as an input when generating $(t+1)$-th output, but it takes lots of training time to overcome errors from model's imperfect output. 
Therefore, we start to train the network with the teacher-forcing method, and slowly increase the ratio of selecting a student-forcing method.

Let $p_{tf}$ denotes the probability to select a teacher-forcing method to train a model. 
For each training step, whether to use the ground truth value or generated value of $\mathbf{p}(t)$ for generating $\mathbf{p}(t+1)$ is selected with a probability of $p_{tf}$.
We start with $p_{tf}=1$, and decay $p_{tf}$ with the factor $\beta_{tf}=0.999$ at every $40,000$ training step. 
In addition, the $l1$-loss function between the ground truth pose vector and estimated pose vector is minimized by the Adam optimizer \cite{Kingma2014Adam:Optimization}, with a learning rate value of $10^{-5}$.

\subsection{Music Genre Classifier}

\begin{figure}[t]
\begin{center}
    \includegraphics[width=\linewidth]{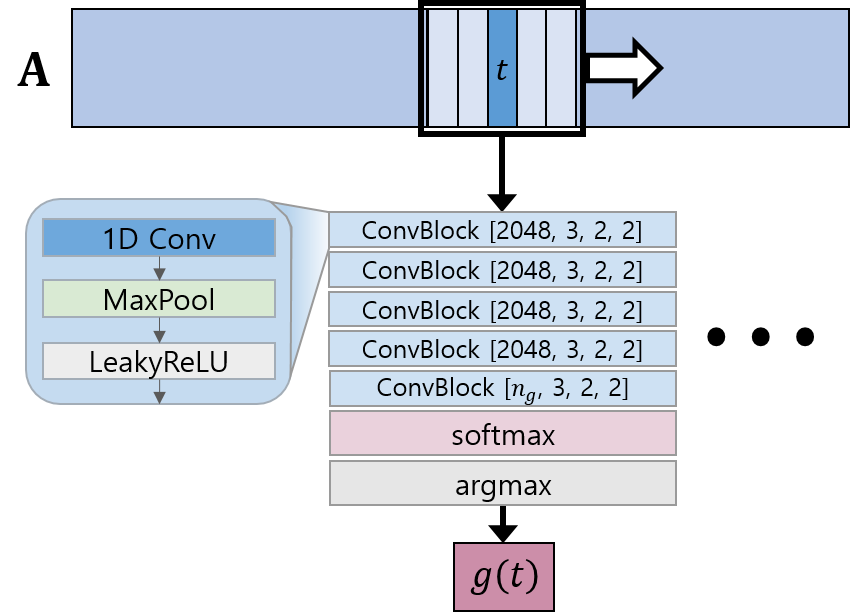}
\end{center}
\caption{The architecture of a music genre classifier. The numbers inside the brackets refer the relevant hyperparameters regarding \texttt{ConvBlock}: the quadruplet indicates the number of output kernels, the size of each kernel, the ratio of subsampling in max-pooling, the stride value in max-pooling.}
\label{fig:classifier}
\end{figure}

\subsubsection{Dataset and Preprocessing}
For training a music genre classifier, 
we also use a dataset in \cite{relate_1},
which provides songs of four music genres, such as Cha-Cha, Rumba, Tango, and Waltz.
The audio signal of each song is converted into a set of audio features $\mathbf{A}=\{\mathbf{a}(0),\ldots,\mathbf{a}(T)\}$,
and used for the training a music genre classifier $\mathcal{C}$.

\subsubsection{Network Architecture}
The proposed music genre classifier $\mathcal{C}$,
takes a set of audio features
$\mathbf{A}(t;w_a)=\{\mathbf{a}(t-w_a/2),\ldots,\mathbf{a}(t+w_a/2)\}$ as inputs,
where the window size $w_a=30$ is used in our experiment.
The structure of the music genre classifier is based on the \texttt{ConvBlock} which is also employed in our pose generator. 
After $\mathbf{A}(t;w_a)$ goes through a series of \texttt{ConvBlocks}, softmax, and argmax layer, the genre estimation value $g(t) \in \{1,\ldots,n_g\}$ is generated as outputs, where $n_g$ is the number of genres in training dataset.
Based on the set of total estimated genre values $\{g(0),\ldots,g(T)\}$, the most estimated genre $\boldsymbol{g}$ is chosen as the final genre of the music. 
Note that we have a set of $n_g$ pose generators for all genres, and $\boldsymbol{g}$-th pose generator is selected to synthesize the dance pose sequence for all time frames.  

\subsubsection{Training}

For training a proposed music genre classifier, 
we use a cross entropy loss function and minimize that loss using the Adam optimizer \cite{Kingma2014Adam:Optimization} with the learning rate of $10^{-4}$ for $6,000$ epochs.
For the validation, the dataset is divided into 56 songs 
consisting of the $126,995$ audio feature vectors for training, 
and six songs consisting of $13,495$ audio feature vectors for testing.
After training, the proposed music genre classifier 
has succeeded in classifying $73.15\%$ of the test audio feature vectors.
However, for classfying the genre of a song, it has succeeded in classifying all of six text data songs correctly.
This is because the genre of the song is determined by taking into account all genre values of the entire song, so even if there are misclassified genre values in a few places, the genre for that song can be correctly classified. 

\begin{figure*}[t]
\begin{center}
    \includegraphics[width=0.95\linewidth]{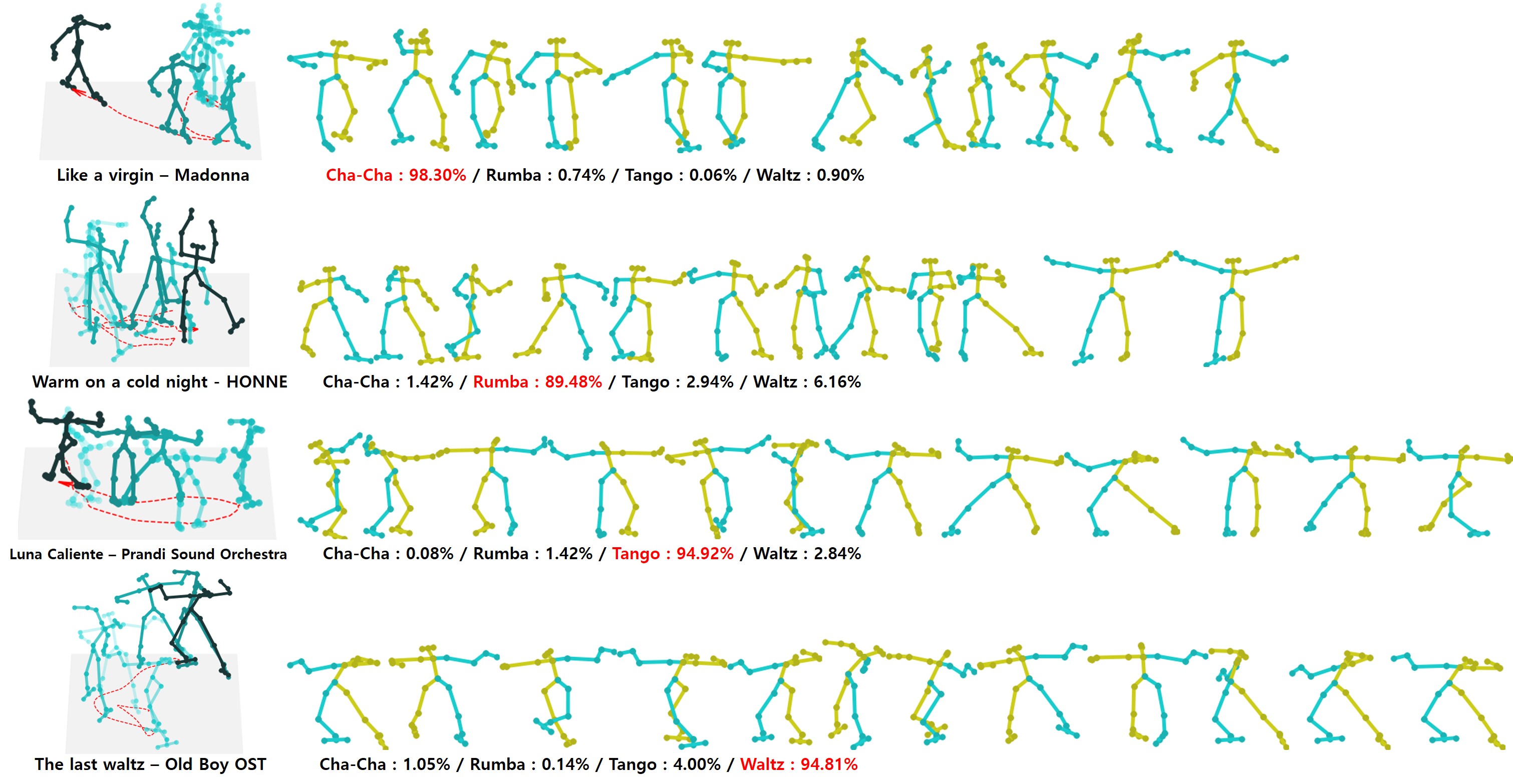}
\end{center}
\caption{Generation results for four music genres. The whole sequence is $12$ seconds. (Left) movement from above. As the time step increases, the color of skeleton darkens. The red line on the ground indicates the path of center of mass.(Right) the corresponding sequence plotted by 1 fps.}
\label{fig:gen_result}
\end{figure*}%
\begin{figure*}[t]
\begin{center}
    \includegraphics[width=0.95\linewidth]{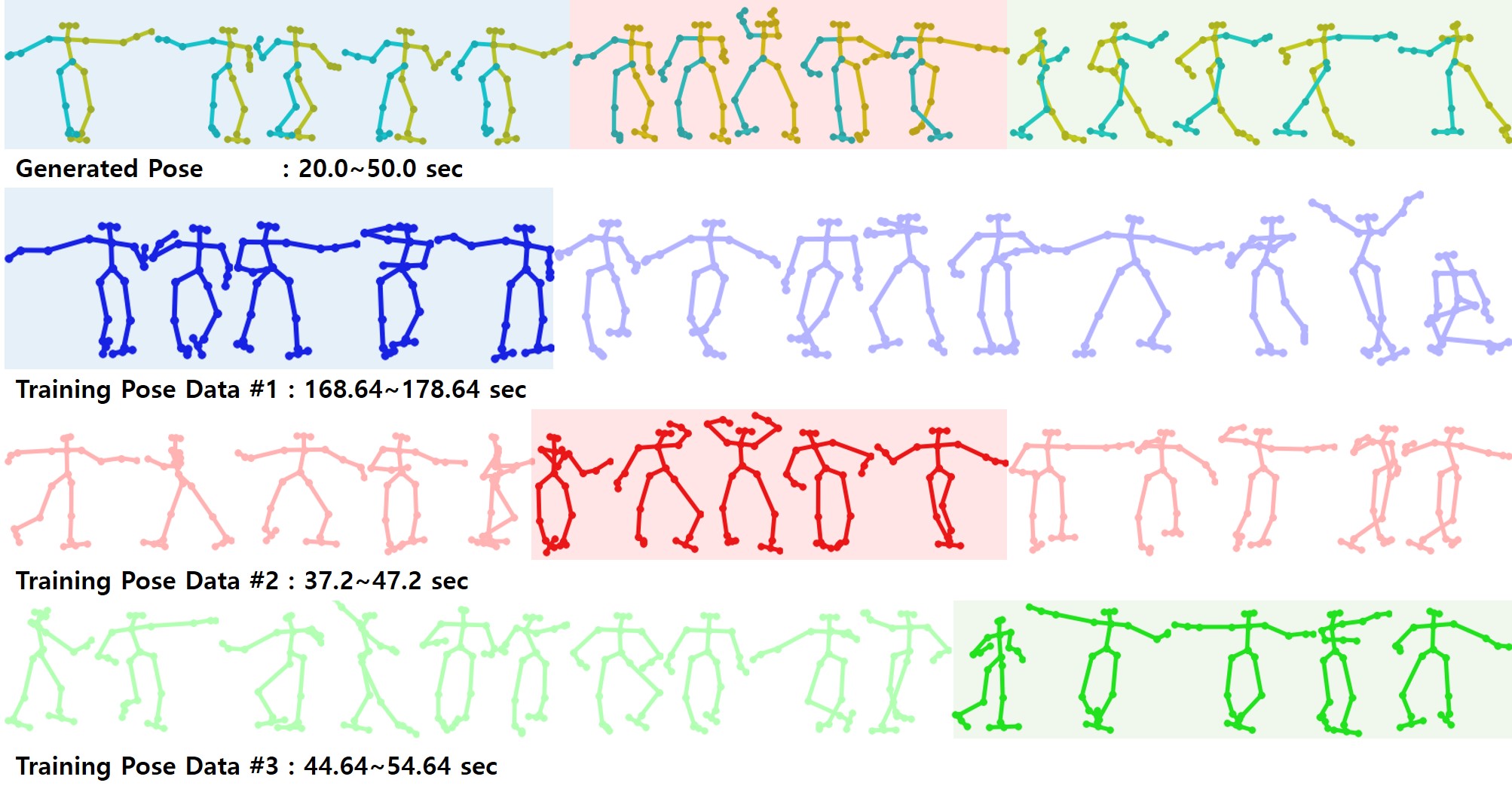}
\end{center}
\caption{The analysis of which part of the training dance data matches to the generated dance. The dance data is plotted with $0.5$ fps, Poses highlighted in the same color are the ones with the highest similarity.}
\label{fig:DTW_result}
\end{figure*}%
\begin{figure}[t]
\begin{center}
    \includegraphics[width=\linewidth]{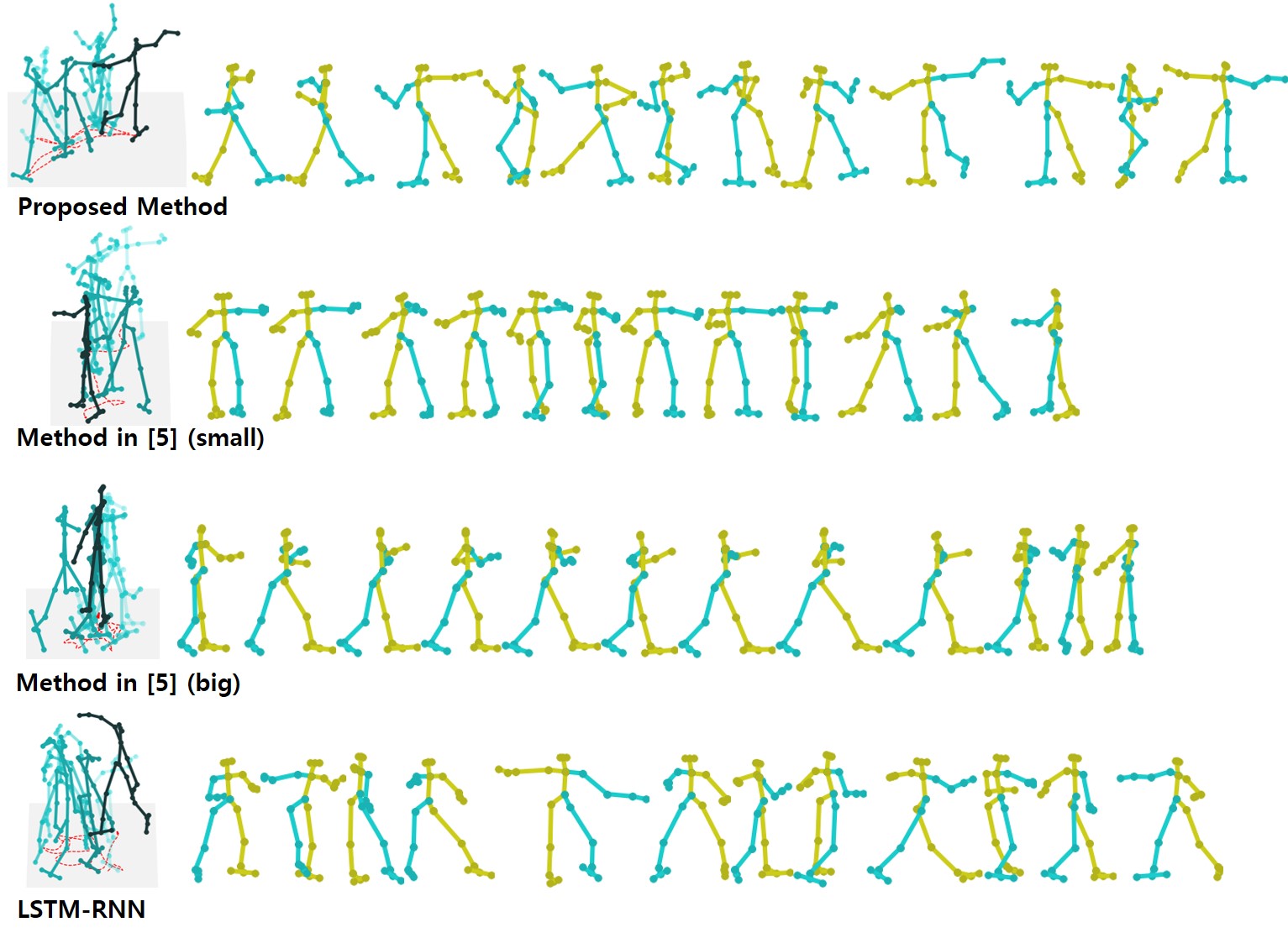}
\end{center}
\caption{Comparison of generation with \cite{relate_1} and baseline method. The whole sequence is $12$ seconds. (Left) movement from above. As the time step increases, the color of skeleton darkens. he red line on the ground indicates the path of center of mass. (Right) the corresponding sequence plotted by 1 fps.}
\label{fig:compare_result}
\end{figure}

\section{Experiments} \label{sec:exp}
\subsection{Dance Generation Results}
Figure~\ref{fig:gen_result} shows the generated dance sequences of $12$ seconds from four songs. 
The result is based on the pose generator selected by the genre classified by the music genre classifier.
These dances are generated from unheard songs and the supplementary video shows related results more vividly.\footnote{Video link : \url{https://youtu.be/AoMUXV3Mubk}}
The supplementary video shows that the proposed framework can generate the movements of a person dancing to the beat of given music, and produce the appropriate dance for the classified genre.

To verify whether the trained pose generator has learned to dance from the training dataset, 
we have analyzed where the generated dance originated from the training data.
After dividing the generated dance and training dance data into a certain time interval, 
we have tried to determine which section of the generated dance is most similar to the training data using dynamic time warping (DTW) \cite{dtw}.
Figure~\ref{fig:DTW_result} shows the analysis of which part of the training dance data matches to the dance generated based on \textit{There's nothing holding me back} by \textit{Shawn Mendes}.
The pose sequences highlighted in the same color are the ones with the highest similarity.
The result shows that the generated dance has a pattern similar to the training data, but its details differ from the training data so that it can match the given input music.
Based on this, we claim that the proposed system has the ability to adapt dance patterns learned from training data to the unheard input music,
and a new dance sequence different from the training dataset can be generated.
%
%
%

\subsection{Comparison with \cite{relate_1} and Baseline Models} \label{sec:exp_compare}

In this section, we compare the proposed pose generator 
with a generative network suggested in \cite{relate_1}.
The network proposed in \cite{relate_1} consists of
an LSTM-based autoencoder
that maps the classic audio features into ones suitable for the dance generation, 
and the LSTM-based generator that synthesizes dance poses based on the generated audio features.
Since no source code for implementing the network is provided,
we have implemented the network according to the instructions described in \cite{relate_1}.
The difference from the original paper is the batch of size 4 has been formed after randomly selecting sections for 250 frames of audio and pose dataset.
In addition, since their method of masking the audio feature does not seem to contribute to the improvement of dance quality, the feature masking part has been excluded.
Also, we have employed the representation of the pose vector used in this paper, so that the center of the generated dance pose can move.

\begin{table}[]
\center
\begin{tabular}{|l|l|l|l|l|}
\hline
      & Proposed & {[}5{]} (Small) & {[}5{]} (Big) & LSTM-RNN \\ \hline
$N_p$ & 4.15M    & 4.17M           & 13.47M        & 4.87M    \\ \hline
\end{tabular}
\caption{Number of network parameters of models in Figure~\ref{fig:compare_result}}
\label{table:np}
\end{table}

Comparison results of our pose generator and the network proposed in~\cite{relate_1} are shown in Figure~\ref{fig:compare_result}. 
In this figure, all dances are generated from the same section of the same song (\textit{Single Ladies - Beyonc\'e}).
From the dance sequences generated by the network proposed in~\cite{relate_1}, 
we have observed many unnatural parts such as the body moving while not walking.
Table~\ref{table:np} shows the number of network parameters of models in Figure~\ref{fig:compare_result}, where $N_p$ denotes the number of network parameters.
Even if the number of parameters of the network proposed in~\cite{relate_1} is set to be similar to or more than that of our network, same result has been observed.
We speculate that the authors of~\cite{relate_1} 
did not observe this phenomenon
because they used the data after the preprocessing procedure 
that keeps the center of the human pose from moving.
For more vivid comparison, refer the supplementary video.

In order to verify the effectiveness of our music feature encoder,
we analyze the results
when the input audio feature for the pose generator of~\cite{relate_1} is the one generated by our music feature encoder.
The result is shown in Figure~\ref{fig:compare_result}
with a label LSTM-RNN.
It is shown that the pose generator of~\cite{relate_1} has been improved when using our high dimensional audio feature, instead of the audio feature encoded from their LSTM-based autoencoder.
However, slight body and foot slidings have been still observed 
even with the number of network parameters similar to our proposed network. (See the supplementary video for details.)

\begin{figure*}[t]
\begin{center}
    \includegraphics[width=\linewidth]{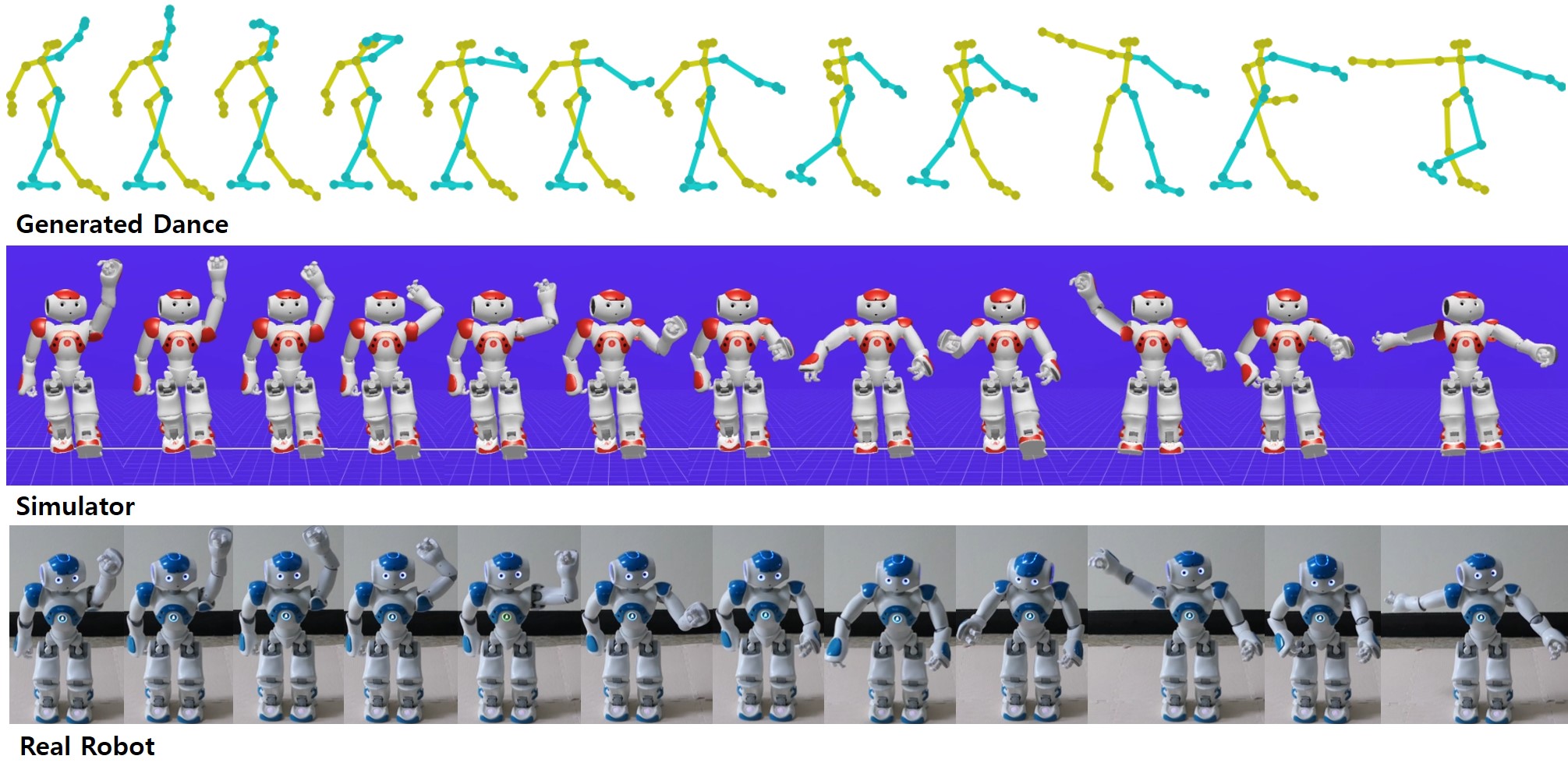}
\end{center}
\caption{Generated dancing moves for song \textit{Dancing is not a crime} by \textit{Panic! at the Disco} (top), a simulated dancing robot (middle), and a real robot dancing to the song (bottom).}
\label{fig:real_exp}
\end{figure*}

\subsection{Dance Generation for a Humanoid Robot NAO}

This section shows the experimental results 
when the dance pose sequence generated by our proposed framework
is applied to a humanoid robot NAO.
To make the humanoid robot follow the generated dance pose sequence, we have solved the inverse kinematics to calculate the joint angle values for controlling the robot. 
The calculated joint angle values are passed to the API provided by NAO, and a path is replanned considering the maximum joint velocity and collision between body parts.
%
However, this replanning process can make the robot's dance different from the original dance, if the movement speed of the dance is too fast.
This is the inevitable problem due to the limitations of the robot hardware. 
To overcome this hardward limitation, a simple remedy is used to generate a video of a dancing robot by making the robot dance slower, shooting the video, and replaying the video faster.
%

%

When the robot dance is implemented on the simulator, 
the full body parts of the robot are moved. 
However, when using the real robot, only the upper body parts were moved since the robot could lose balance and fall.
Figure~\ref{fig:real_exp} shows the result of transferring the generated Rumba dance to the robot (Music: \textit{Dancing is not a crime} by \textit{Panic! at the Disco}).
In this figure, the results of 10 second dance are shown based on 1 fps.
This dance was not fast enough to be implemented on the robot at the original speed.
For more vivid results, please refer the supplementary video.

\section{Conclusion}
In this paper, we have proposed a machine learning based framework for synthesizing a 3D dance pose sequence of a human when a music has been given as an input.
The proposed framework consists of three parts:
a music feature encoder, a pose generator, and a music genre classifier. 
From a given input music, a music feature encoder extracts a set of audio features. 
Based on this, the genre of the music is determined by the music genre classifier, and a pose generator trained for that genre is used to generate the dance pose sequence for all frames.
The proposed pose generator is a generative autoregressive model, 
which takes the current output pose as an input for generating the next pose frame.

The disadvantage of the proposed method is that the pose generator must be trained separately for each genre.
If we trained all genres of dance so that one model could learn, 
it has been observed that the movements unrealistic dancing moves are generated.
In order to construct a model that can learn patterns of various genres of dance, it will be necessary to apply a multi-task learning technique, which is our future work.




\bibliography{curr_bib}
\bibliographystyle{IEEEtran}
\end{document}